\documentclass[runningheads]{llncs}

\makeatletter
\def\@fnsymbol#1{\ensuremath{\ifcase#1\or *\or \dagger\or \ddagger\or
   \mathsection\or \mathparagraph\or \|\or **\or \dagger\dagger
   \or \ddagger\ddagger \else\@ctrerr\fi}}
\makeatother

\usepackage{graphicx}

\usepackage{tikz}
\usepackage{comment}
\usepackage{amsmath,amssymb} 
\usepackage{color}

\usepackage[accsupp]{axessibility}  

\usepackage{multirow}
\usepackage{tabularx}
\usepackage{booktabs}
\usepackage{subcaption}
\usepackage{orcidlink}
\newcolumntype{C}{>{\centering\arraybackslash}X}

\newcommand{\eg}{\emph{e.g.}}
\newcommand{\ie}{\emph{i.e.}}

\newcommand{\spaceline}{\ -\ }

\newcommand{\grey}[1]{\textcolor[rgb]{0.745, 0.745, 0.745}{#1}}

\begin{document}
\pagestyle{headings}
\mainmatter
\def\ECCVSubNumber{7562}  

\title{Motion Sensitive Contrastive Learning for Self-supervised Video Representation} 

\titlerunning{Motion Sensitive Contrastive Learning}
%
\author{Jingcheng Ni\inst{1,2}\orcidlink{0000-0001-6276-6296} \and
Nan Zhou\inst{1,2}\orcidlink{0000-0003-3443-6171} \and
Jie Qin\inst{3}\orcidlink{0000-0002-0306-534X}$^{\ast}$ \and
Qian Wu\inst{4}\orcidlink{0000-0002-4032-7169} \and\\
Junqi Liu\inst{4}\orcidlink{0000-0002-4801-5282} \and
Boxun Li\inst{4}\orcidlink{0000-0002-6132-7547} \and
Di Huang\inst{1,2}\orcidlink{0000-0002-2412-9330}\thanks{Corresponding authors.}
}
\authorrunning{J. Ni et al.}
%
\institute{
State Key Laboratory of Software Development Environment, \\Beihang University, Beijing, China  \and
School of Computer Science and Engineering, Beihang University, Beijing, China \and
College of Computer Science and Technology, \\Nanjing University of Aeronautics and Astronautics, Nanjing, China \and 
MEGVII Technology}

\maketitle

\begin{abstract}

Contrastive learning has shown great potential in video representation learning. However, existing approaches fail to sufficiently exploit short-term motion dynamics, which are crucial to various downstream video understanding tasks. In this paper, we propose Motion Sensitive Contrastive Learning (MSCL) that injects the motion information captured by optical flows into RGB frames to strengthen feature learning. To achieve this, in addition to clip-level global contrastive learning, we develop Local Motion Contrastive Learning (LMCL) with frame-level contrastive objectives across the two modalities. Moreover, we introduce Flow Rotation Augmentation (FRA) to generate extra motion-shuffled negative samples and Motion Differential Sampling (MDS) to accurately screen training samples. Extensive experiments on standard benchmarks validate the effectiveness of the proposed method. With the commonly-used 3D ResNet-18 as the backbone, we achieve the top-1 accuracies of 91.5\% on UCF101 and 50.3\% on Something-Something v2 for video classification, and a 65.6\% Top-1 Recall on UCF101 for video retrieval, notably improving the state of the art.

\end{abstract}

\section{Introduction}

Video understanding has become a necessity in the past decade due to the rapid and massive growth of data. In this challenging task, video representation is the most fundamental and important issue and has received consistently increasing attention. In the literature, many efforts have been made along with the release of several large-scale benchmarks, such as Kinetics \cite{17arxiv/kinetics} and YouTube-8M \cite{16arxiv/youtube}, where representations are learned in a supervised manner from manually annotated samples. Unfortunately, building such databases inevitably incurs enormous human and time cost.

Self-supervised learning has recently emerged as a promising alternative in visual representation. Different from the case on images that only considers spatial variations, that on videos puts more emphasis in temporal characteristics. A number of studies on videos have shown huge potential to learn general features by making use of a tremendous amount of unlabeled data available on the Internet, facilitating diverse downstream applications, including action recognition, action detection, video retrieval, \emph{etc}.

Among current self-supervised video representation learning methods, contrastive learning based ones \cite{21cvpr/largescale,21cvpr/cvrl} have delivered a great success. They treat the clips from the same video as positive pairs while the ones from different videos as negative pairs and apply the InfoNCE loss \cite{18arxiv/infonce} to train the model, which is expected to distinguish the clips of a given video from the ones of others. However, clip-level contrastive learning is relatively coarse and primarily benefits global (\emph{a.k.a.} long-term) features \cite{extra/slowfeature} without meticulously capturing local (\emph{a.k.a.} short-term) dynamics, thus limiting the performance, in particular in fine-grained scenarios.

\begin{figure}[!t]
    \centering
    \includegraphics[width=0.99\textwidth]{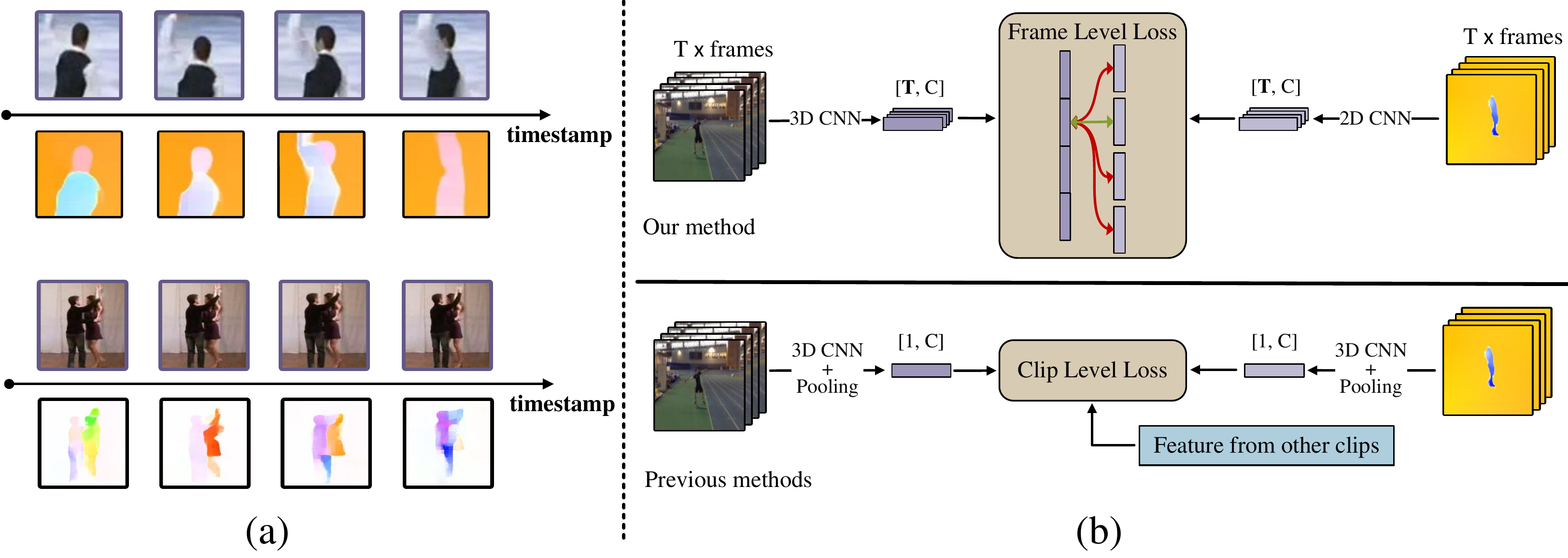}
    \caption{(a) The RGB frames are not sensitive enough to short-term motion changes, while the optical flows are able to capture subtle motion dynamics between frames, where the changes of motion vectors (in different colors) are clearly observed from the flow maps. (b) Comparison between our method and previous ones using optical flows. Existing works generate clip-level features with temporal pooling, while we focus on more fine-grained frame-level features.}
    \label{Fig:1}
\end{figure}

More recently, some attempts have compensated this by designing and conducting contrastive learning across video clips with additional views, \emph{e.g.} global \emph{vs.} local \cite{21arxiv/tclr} and long \emph{vs.} short \cite{21iccv/lsfd}. Although local temporal modeling is enhanced to some extent with decent improvements reported, they still suffer from two major downsides. On the one hand, for the continuity and redundancy of video data, it is really difficult to handle the discrepancy between the frames within a small time slot, \emph{e.g.} at adjacent timestamps, without high-level supervision, making their representations not sufficiently powerful. On the other hand, existing global-local or long-short contrastive learning requires repetitive temporal interval sampling, leading to multiple forward processes, for a single video, which is both time- and memory-consuming.

In this paper, we propose a novel self-supervised contrastive based approach for video representation learning, namely Motion Sensitive Contrast Learning (MSCL). To overcome the shortcomings aforementioned, besides encoding the global motion from RGB frames, it also fully exploits local temporal clues by introducing optical flows since they prove sensitive to very short-term dynamics, as illustrated in Fig. \ref{Fig:1} (a).
To fulfill this, we propose Local Motion Contrastive Learning (LMCL) that directly leverages optical flows as the supervisory signal for frame-level local dynamics learning. Specifically, LMCL matches cross-modality (RGB \emph{vs.} optical flows) features at the same timestamp so that subtle motions are modeled. Meanwhile, to restrict the temporal receptive field of flow features in frame-level contrast, different from previous works \cite{21iccv/broaden,20nips/coclr,21arxiv/modist} on clip-level contrast, we adopt a lightweight 2D CNN as the encoder without temporal message passing, as illustrated in Fig. \ref{Fig:1} (b) and elaborated in Section \ref{ofvu}. In this way, those features capturing local dynamics can be efficiently obtained from the 2D flow encoder, bypassing the cumbersome phase of extra local interval sampling required in \cite{21iccv/lsfd,21arxiv/tclr}.
In addition, we present two practical strategies to further facilitate LMCL. First, as LMCL introduces frame-level contrast, clips with limited motions tend to bring negative effects to the learning process. To solve this problem, we design Motion Differential Sampling (MDS) to enhance the sampling probability of clips with large motion differential. Second, Flow Rotation Augmentation (FRA) takes rotated flows with different motion vectors as extra negative samples, thereby highlighting motion information on local features.

We summarize our main contributions as follows:
\begin{itemize}
    \item We propose LMCL, taking advantage of optical flows to underline subtle motions for frame-level contrastive learning, which substantially strengthens self-supervised video representations.
    \item We present MDS and FRA to optimize temporal interval sampling and optical flow augmentation respectively, both of which further facilitate LMCL.
    \item We achieve competitive results on several standard benchmarks, \emph{i.e.}, UCF101, HMDB51, and Something-Something v2, in video classification and retrieval.
\end{itemize}

\section{Related Work}

\subsection{Self-supervised Video Representation Learning}
Various self-supervised video representation learning methods have been devised to take advantage of unlabeled video data on the Internet.
These methods learn to accomplish various human-designed pre-tasks, including frame sorting \cite{17iccv/urlss}, pace prediction \cite{20eccv/sslpp}, speed prediction \cite{20cvpr/speednet,21cvpr/staticimages}, and spatio-temporal jigsaw solving \cite{21ijcai/csj,extra/videojigsaw}.
More recent works have been inspired by the success of contrastive learning in the image domain such as \cite{20cvpr/mocov1,20icml/simclr,20nips/byol,20nips/swav}, which can be viewed as instance discrimination tasks \cite{18cvpr/instdisc}.
Since videos contain extra attributes that contribute to distinguishing instances, \cite{21cvpr/cvrl,21cvpr/largescale} take time-shift as the invariant attribute and \cite{21iccv/ascnet,21aaai/rspnet} regard speed as the variant attribute. More generally, multiple attributes are explored jointly by the combination of temporal transforms in \cite{21iccv/te,21iccv/compositions}. In \cite{21iccv/space}, transforms are performed in the feature space to reduce memory consumption.
%
\subsection{Optical Flows in Video Understanding}
\label{ofvu}
Temporal information is of high importance in video understanding. Optical flows, corresponding to motion vectors across frames, have shown potential in modeling dynamics in the two-stream structure \cite{14nips/twostream}. However, \cite{extra/integration,13iccv/understanding} indicate that motion information does not work as expected. For example, in supervised action recognition, the shuffling operation at input stage has much more impact on RGB sequences than flow ones. The motivation behind is on the property of appearance-invariance \cite{extra/integration}, where the motion foreground is described with a low variety. 
Some self-supervised learning methods can also be seen as taking advantage of this property. COCLR \cite{20nips/coclr} highlights the prior that videos belonging to the same class have similar flow patterns but significant variations in the RGB space, and MoDist \cite{21arxiv/modist} distills the flow information to make RGB features focus more on motion foreground. Unlike these methods, we make use of the motion information itself as guidance to improve the ability of modeling local dynamics in RGB features.

\subsection{Fine-grained Temporal Features}
Learning local dynamics is an important topic for video understanding. Some works attempt to improve modeling through dedicatedly designed structures \cite{21cvpr/tdn,19iccv/stm} or explicit constraints \cite{20eccv/tdrl}. But in the self-supervised learning domain, existing methods do not pay enough attention to fine-grained temporal features. For instance, \cite{21cvpr/largescale,21cvpr/cvrl} can be viewed as learning slow features \cite{extra/slowfeature}, which are more relevant to scene information. To encode more temporal clues in self-supervised video representation learning, LSFD \cite{21iccv/lsfd} introduces feature contrast between long-term and grouped short-term features, and TCLR \cite{21arxiv/tclr} directly compares features with different time-spans. Although these works do make improvements, they require extra sampled clips in the forward process for short views and cannot take advantage of local motion information in flows. 

\begin{figure}[t!]
    \centering
    \includegraphics[width=0.99\textwidth]{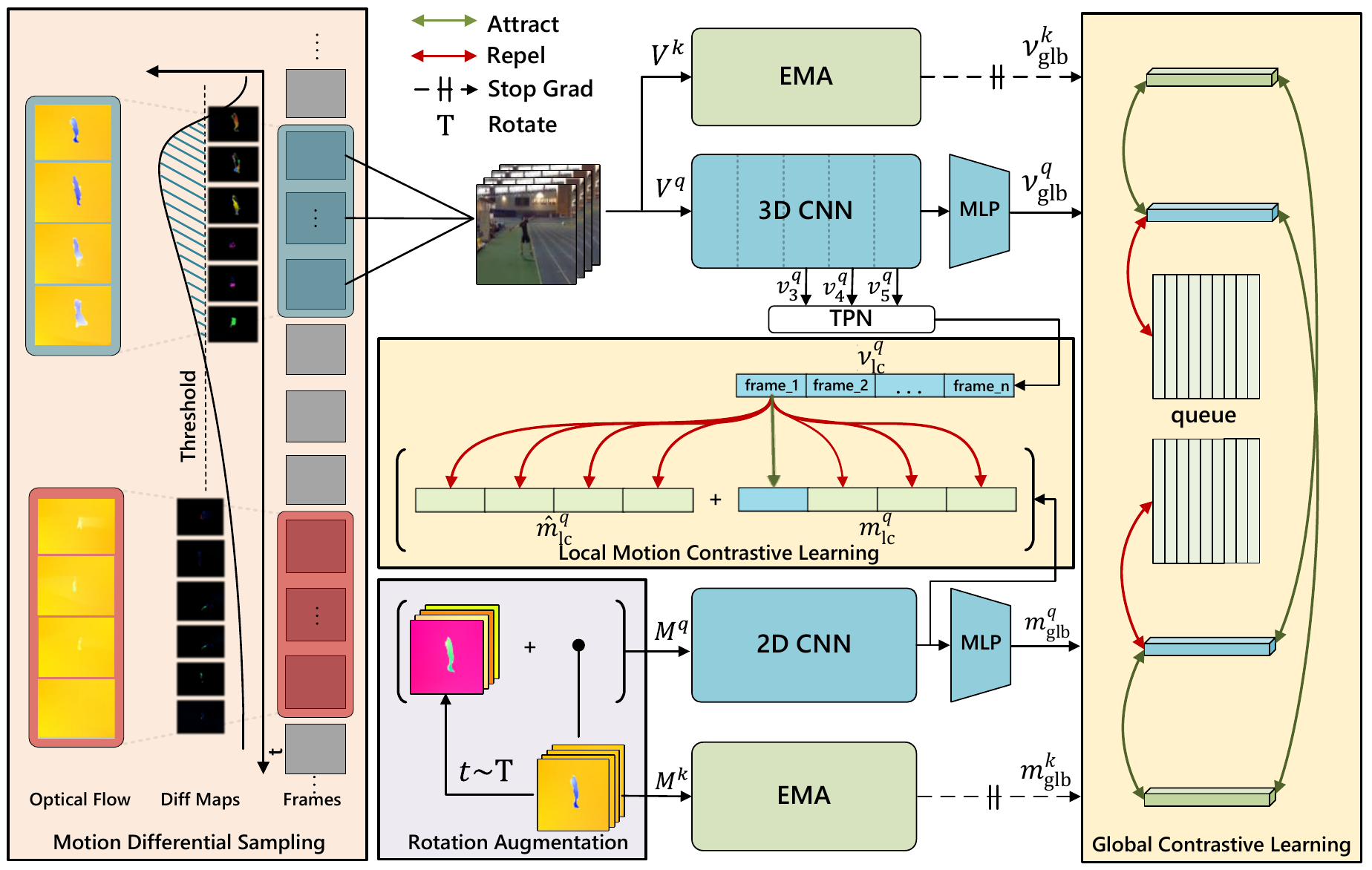}
    \caption{Overview of the proposed Motion Sensitive Contrast Learning (MSCL) framework. Clips with large motion differentials are sampled from the video and random augmentations are employed for generating queries $\{V^{q}, M^{q} \}$ and keys $\{V^{k}, M^{k} \}$ w.r.t. the RGB and flow modalities. Then, global features $\{ \nu^q_{\mathrm{glb}},\nu^k_{\mathrm{glb}},m^q_{\mathrm{glb}},m^k_{\mathrm{glb}} \}$ are extracted for global contrastive learning, where EMA indicates the momentum key encoder in \cite{20cvpr/mocov1}. In order to inject motion dynamics from flow to RGB features,  Local Motion Contrastive Learning (LMCL) is conducted based on local features $\{ \nu^q_{\mathrm{lc}},m^q_{\mathrm{lc}}\}$. To further enforce the model to focus on the motion information, rotation augmentation is applied on the flow inputs and the corresponding local features $\hat{m}^q_{\mathrm{lc}}$ are used to enhance negative samples in LMCL.}
    \label{fig:2}
\end{figure}

\section{Method}
\label{sec:blind}
In this section, we introduce the proposed self-supervised framework for video representation learning in detail. We begin by introducing the global and local feature extraction pipelines for the two modalities (\ie, RGB and optical flow), respectively. Subsequently, we present the commonly-used global contrastive losses and the proposed Local Motion Contrastive Learning (LMCL), including sample pair construction and augmentation. Finally, we introduce the motion differential sampling policy, which provides meaningful samples for learning more effective local temporal features. An overview of our proposed framework is illustrated in Figure \ref{fig:2}.

\subsection{Global and Local Feature Extraction}
Given a sequence of video frames, we first extract optical flow images from pairs of frames with stride $s$. We learn both clip-level global and frame-level local features in terms of both RGB and optical flow modalities. Specifically, we adopt the spirit of contrastive learning to build self-supervised features, and randomly sample video clips $\{V^{q}, V^{k} \}$ with the corresponding optical flows $\{M^{q}, M^{k} \}$ as the queries and keys. 
As we follow the symmetric structure of MoCo \cite{20cvpr/mocov1} where queries and keys are encoded in a similar way, we only elaborate how to extract features for queries for brevity.

In the RGB pathway, a 3D CNN is employed as the feature encoder, where the output features of different stages (\ie, conv3, conv4 and conv5 layers) are denoted as $\{\nu^q_3, \nu^q_4, \nu^q_5 \}$, respectively. 
We extract video-level global features based on the output of the last stage (\ie, $\nu^q_5$). More concretely, we apply spatio-temporal pooling on $\nu^q_5$, followed by a 2-layer MLP as the projection head. For frame-level local features, we use a 3-layer Temporal Pyramid Network (TPN) \cite{20cvpr/tpn} to merge multi-level features. Specifically, we compute the global features $\nu^q_{\mathrm{glb}}$ and the local features $\nu^q_{\mathrm{lc}}$ in the RGB pathway as follows:
\begin{equation}\label{(1)}
   \begin{split}
      & \nu^q_{\mathrm{glb}} = \textrm{MLP}(\textrm{STPool}(\nu^q_5)), \\
      & \nu^q_{\mathrm{lc}} = \textrm{SPool}(\hat{\nu}^{q}_{3}),~\hat{\nu}^q_{3}, \hat{\nu}^q_{4}, \hat{\nu}^q_{5} = \textrm{TPN}(\nu^q_3, \nu^q_4, \nu^q_5),
   \end{split}
\end{equation}
where $\textrm{STPool}(\cdot)$ and $\textrm{SPool}(\cdot)$ indicate spatio-temporal pooling and spatial pooling, respectively. In practice, we
only use the first-stage output $\hat{\nu}_3^q$ from TPN,
due to two reasons. First, the receptive fields of temporal features in later stages of 3D CNNs (especially those with temporal down-sampling, \eg, S3D \cite{17cvpr/s3d} and R(2+1)D \cite{18cvpr/r21d}) are too large to compute the frame-level contrastive loss. Second, TPN makes local features benefit more from multi-level information conveyed in the RGB image.

For the flow pathway, we use a 2D CNN as the feature encoder to extract optical flow features. 
This design pushes the feature only containing the temporal information in the single flow which benefit LMCL in Section \ref{lmcl}, and avoid the temporal position leakage \cite{20arxiv/position} by zero padding in 3D CNN. Since the variability of flows is less than that of RGB frames, we decrease the number of channels to 1/8 of that of the RGB counterpart, as suggested in \cite{19iccv/slowfast,21arxiv/modist}. Different from the RGB pathway, there exists no temporal down-sampling (\ie, temporal receptive field is not enlarged), so we directly apply the output of the last stage $m^q_5$ to extract both global and local features as follows:
\begin{equation}\label{(2)}
   \begin{split}
      & m^q_{\mathrm{glb}} = \textrm{MLP}(\textrm{STPool}(m^q_5)), \\
      & m^q_{\mathrm{lc}} = \textrm{SPool}(m^{q}_{5}).
   \end{split}
\end{equation}
Similarly, we can obtain the global features of the keys $V^{k}$ and $M^{k}$ w.r.t. RGB and flow as $\nu^k_{\mathrm{glb}}$ and $m^k_{\mathrm{glb}}$, respectively. Note that there is no need to extract local features for the keys as the local contrastive learning is applied on the query features only (see Sec. \ref{lmcl}).

\subsection{Global Contrastive Learning}
In this section, we introduce the global contrastive learning based on clip-level features. We follow the basic pipeline of MoCo \cite{20cvpr/mocov1} for both pathways, where the momentum encoder is employed for key inputs, and the memory bank is used for saving negative clips. There are two types of global contrastive losses in our method. The first intra-modality loss is applied on the features from either the RGB or the flow modality to make them discriminative in their own domains. Specifically, we adopt the widely-used InfoNCE \cite{18arxiv/infonce} loss for global contrastive learning, which is defined as:
\begin{equation}\label{(3)}
   \begin{split}
      & L_\mathrm{RGB}=-log\frac{h(\nu^q_{\mathrm{glb}}, \nu^k_{\mathrm{glb}})}{h(\nu^q_{\mathrm{glb}}, \nu^k_{\mathrm{glb}})+\sum_{i=1}^{N}h(\nu^q_{\mathrm{glb}}, \bar{\nu}_{i,\mathrm{glb}}^k)}, \\
      & L_\mathrm{Flow}=-log\frac{h(m^q_{\mathrm{glb}}, m^k_{\mathrm{glb}})}{h(m^q_{\mathrm{glb}}, m^k_{\mathrm{glb}})+\sum_{i=1}^{N}h(m^q_{\mathrm{glb}}, \bar{m}_{i,\mathrm{glb}}^k)},
   \end{split}
\end{equation}
where $h(x, y)=\textrm{exp}(x^Ty / \|x\| \|y\| \tau)$ is the distance between two feature vectors $x$ and $y$; $\bar{\nu}_{i,\mathrm{glb}}^k$ and $\bar{m}_{i,\mathrm{glb}}^k$ represent the $i$-th global features of the two modalities in the memory bank with size $N$, respectively;
$\tau$ is the temperature parameter. The second inter-modality loss is applied across the two modalities, with the aim of making the RGB features focus more on motion foreground areas \cite{21arxiv/modist}, which contribute to the subsequent local contrastive learning. Concretely, the loss term is formulated as follows:
\begin{equation}\label{(4)}
   \begin{split}
      L_{\mathrm{RF}}=&-(log\frac{h(\nu^q_{\mathrm{glb}}, m^k_{\mathrm{glb}})}{h(\nu^q_{\mathrm{glb}}, m^k_{\mathrm{glb}})+\sum_{i=1}^{N}h(\nu^q_{\mathrm{glb}}, \bar{m}_{i,\mathrm{glb}}^k)} +\\
      &log\frac{h(m^q_{\mathrm{glb}}, \nu^k_{\mathrm{glb}})}{h(m^q_{\mathrm{glb}}, \nu^k_{\mathrm{glb}})+\sum_{i=1}^{N}h(m^q_{\mathrm{glb}}, \bar{\nu}_{i,\mathrm{glb}}^k)}) \ , \\
   \end{split}
\end{equation}
\noindent and different from \cite{21arxiv/modist}, the positive and negative keys come from the same modality, which we find in our experiments is more stable in early training. With the intra- and inter-modality losses, we can obtain discriminative global features for both modalities, and at the same time focus on motion foreground areas, which are essential for the subsequent local motion contrastive learning  phase.

\subsection{Local Motion Contrastive Learning}
\label{lmcl}
The features learned based on the global contrastive losses have difficulty in modeling local dynamics. To address this, we use optical flows to capture local motion information, and introduce frame-level contrastive losses for learning time-variant features. For the local features $\nu^q_{\mathrm{lc}}$ and $m^q_{\mathrm{lc}}$,
we only take frame-level features at the same timestamp as positive pairs. Thus, the local motion contrastive loss can be formulated as:
\begin{equation}\label{(5)}
   \begin{split}
      & L_{\mathrm{LMC}}=-\sum_{i=1}^{T}log\frac{h(\nu^q_{\mathrm{lc}}(i), m^q_{\mathrm{lc}}(i))}{h(\nu^q_{\mathrm{lc}}(i), m^q_{\mathrm{lc}}(i))+\sum_{j=1, j\ne i}^{T}h(\nu^q_{\mathrm{lc}}(i), m^q_{\mathrm{lc}}(j))} \ ,\\
   \end{split}
\end{equation}
where $\nu^q_{\mathrm{lc}}(i)$ represents the frame-level features at timestamp $i$, and $T$ is the total number of sampled video frames. 

\begin{figure}[!t]
  \centering
  \includegraphics[scale=0.26]{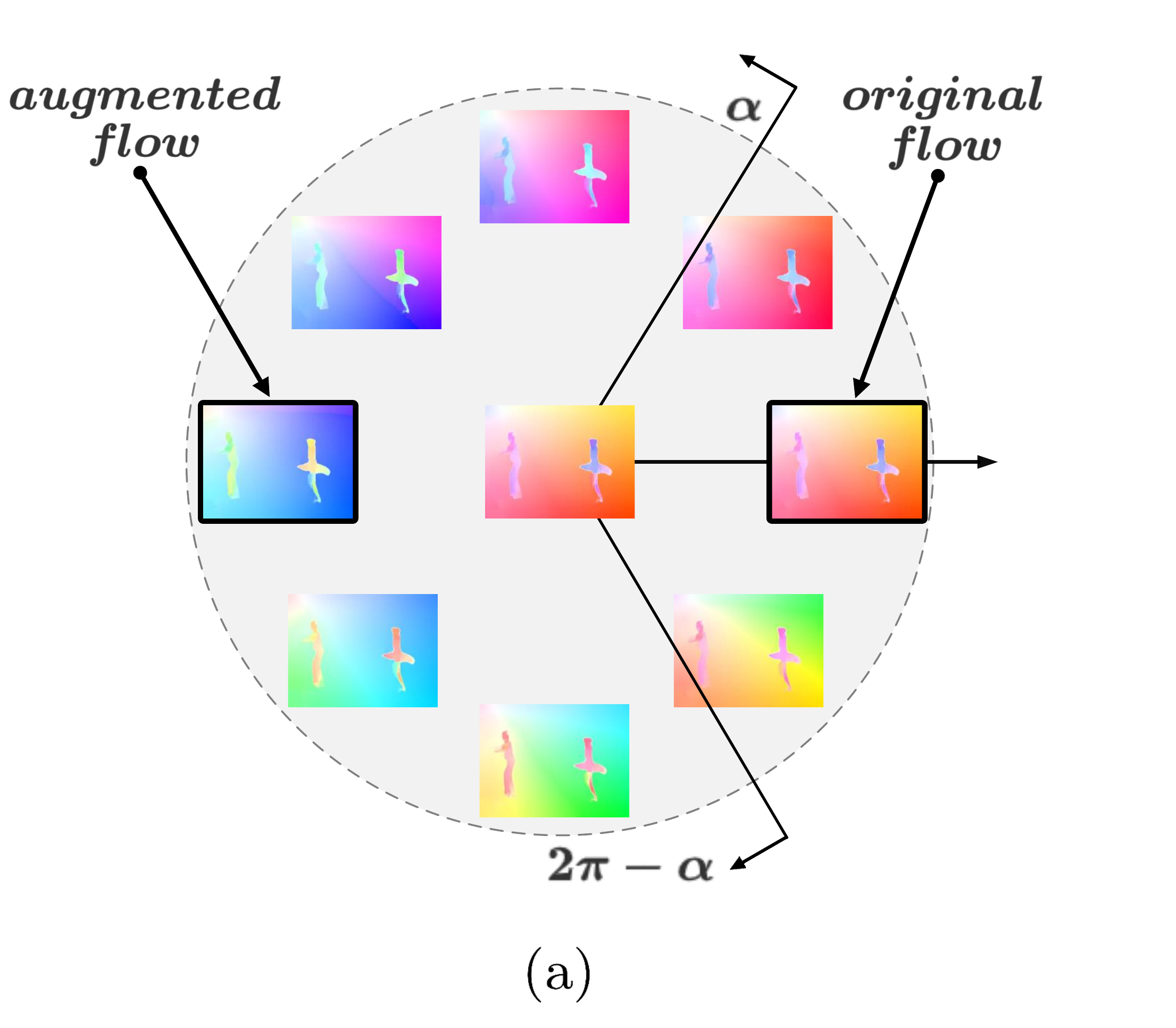}
  \hspace{1em}
  \includegraphics[scale=0.25]{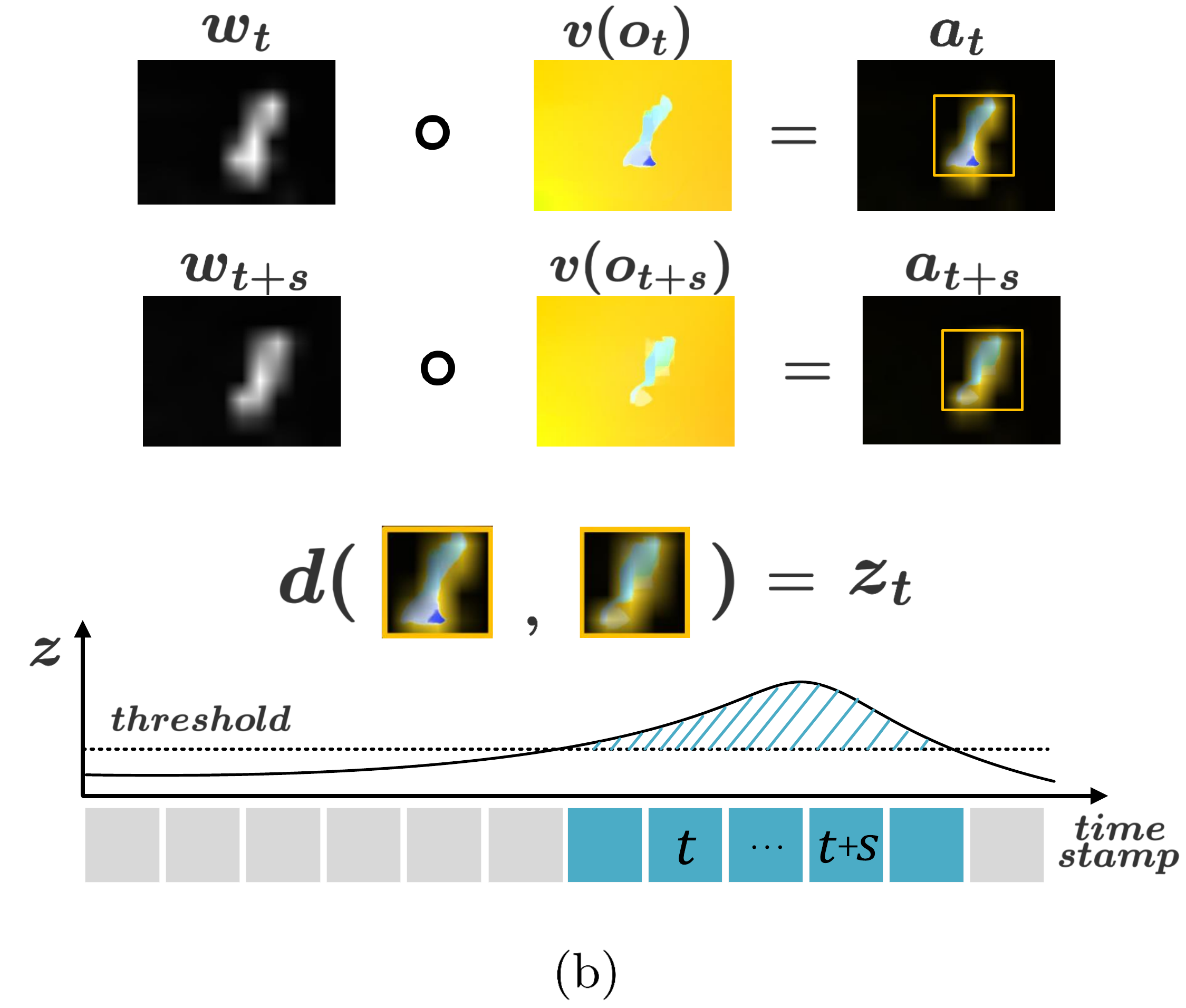}
  \caption{(a) Illustration of Flow Rotation Augmentation (FRA). We randomly sample one angle from $[\alpha, 2\pi-\alpha]$ and rotate the motion vector. (b) Illustration of Motion Differential Sampling (MDS). The notations are consistent with those in Eq. \eqref{(9)}. The foreground regions are highlighted for better viewing.}
  \label{fig:3}
\end{figure}

\noindent \textbf{Flow Rotation Augmentation}. To make this loss focus more on local motion information, we augment the optical flow with extra negative samples. Specifically, the flow is rotated with a specific angle that is randomly sampled from $[\alpha, 2\pi-\alpha]$ and the corresponding local features after rotation are treated as negative samples. In this manner, the local motion contrastive loss turns into:
\begin{equation}\label{(6)}
   \begin{split}
      & A(i)=\sum_{j=1, j\ne i}^{T}h(\nu^q_{\mathrm{lc}}(i), m^q_{\mathrm{lc}}(j))+\sum_{j=1}^{T}h(\nu^q_{\mathrm{lc}}(i), \hat{m}^q_{\mathrm{lc}}(j))\ , \\
      & L_{\mathrm{LMC}}'=-\sum_{i=1}^{T}log\frac{h(\nu^q_{\mathrm{lc}}(i), m^q_{\mathrm{lc}}(i))}{h(\nu^q_{\mathrm{lc}}(i), m^q_{\mathrm{lc}}(i))+A(i)} \ ,\\
   \end{split}
\end{equation}
where $\hat{m}^q_{\mathrm{lc}}$ is the local query feature of the augmented flow. As shown in Figure \ref{fig:3} (a), the augmented flow shares the same outline but different motion vectors (reflected in different colors) with the original one. The augmentation strategy improves the ability of distinguishing local flows by motion vectors, which is in line with the goal of learning local dynamics.

The final learning objective is a linear combination of the global and local contrastive losses, simply weighted by a hyperparameter $\lambda$:
\begin{equation}\label{(7)}
   \begin{split}
      L = L_\mathrm{RGB}+L_\mathrm{Flow}+L_\mathrm{RF}+\lambda L_\mathrm{LMC}'.
   \end{split}
\end{equation}

\subsection{Motion Differential Sampling}
Temporal distinct features are expected to be learned after optimizing the above LMC loss. However, if the motion is similar across different timestamps, it is still difficult for the model to distinguish the corresponding local features. To address this issue, Motion Differential Sampling (MDS) is proposed to enhance the training procedure by choosing samples with larger motion differentials in the foreground. More concretely, suppose $o_t=\{x_t, y_t\}$ and $o_{t+s}=\{x_{t+s}, y_{t+s}\}$ are two adjacent flow maps, we calculate the weighted map $w_t$ that coarsely locates the motion foreground at timestamp $t$ as:
\begin{equation}\label{(8)}
   \begin{split}
      & w_t=\textrm{softmax}(\textrm{up\_down}(\textrm{Sobel}(o_t))), \\
   \end{split}
\end{equation}
where $\textrm{Sobel}(\cdot)$ is the Sobel operator \cite{extra/sobel} for motion boundary detection and $\textrm{up\_down}(\cdot)$ is a coarsening operator implemented by downsampling and upsampling with stride $r$, which is set to 28 in practice. Then, the motion differentials in the foreground area can be defined as:
\begin{equation}\label{(9)}
   \begin{split}
      & a_t = v(o_t) \circ w_t,\ a_{t+s} = v(o_{t+s}) \circ w_{t+s},\\
      & z_t=\textrm{sum}(d(a_t, a_{t+s})) \ ,\\
   \end{split}
\end{equation}
where $v(\cdot)$ follows \cite{extra/evaluation} to convert flows into RGB images, $d(\cdot)$ calculates the Euclidean distance along the channel dimension, $s$ is the stride between frames, $\circ$ is the Hadamard product, and $\textrm{sum}(\cdot)$ is the summation operation in the spatial domain. More concretely, at timestamp $t$, we first calculate the pixel-level distance map between two masked flow maps in the RGB space, and then the differential value $z_t$ is the summation of the distance map. As the samples with larger motion differentials in the foreground contribute more to LMCL, we use the differential value as measurement. For one clip, we take the averaged frame-level differential values as the clip-level score. Finally, MDS is performed by choosing those clips, whose differential values are above the threshold, which is simply the median of the values of all candidate clips in the same video.




\section{Experiments}

\subsection{Datasets}
The UCF101 \cite{12arxiv/ucf101} and Kinetics400 (K400) \cite{17arxiv/kinetics} datasets are used for pre-training. To evaluate the action classification task, we conduct experiments on UCF101, HMDB51 \cite{iccv11/hmdb}, as well as Something-Something v2 (SSv2) \cite{17iccv/ssv2}. As for the video retrieval task, we employ the UCF101 and HMDB51 datasets.

\subsection{Experimental Settings}
{\bf Backbones.} For the RGB pathway, we follow the common practice in \cite{21iccv/lsfd,21arxiv/tclr} and choose ResNet3D-18 (R3D-18) as the general backbone. We also use S3D \cite{17cvpr/s3d} as the auxiliary backbone for apple-to-apple comparison with more counterparts. 
For the flow modality, thanks to its simple appearances, we always use ResNet-18 with 1/8 channels as the backbone.

{\noindent\bf Pre-training Details.} The input clip contains 8 frames with the stride of 16 for K400 and 8 for UCF101, as the latter does not have sufficient frames for large strides. We use a 2-layer MLP like \cite{20arxiv/mocov2} for both pathways. For the RGB inputs, we follow the augmentation in \cite{20arxiv/mocov2} including random grayscale, color jitter, Gaussian blur and horizontal flip. For the motion inputs, we extract flows by RAFT \cite{20eccv/raft} and visualize them to 3-channel images. For the consistency in motion information, we copy the flip transforms in RGB inputs to the corresponding motion ones and ignore other augmentations. During training, $\alpha$ is set to $\pi/3$ and the rotation angle keeps consistent in one clip. We set the memory size $N$ to 65,536 and the temperature $\tau$ is 0.07. The model is pre-trained with 200 epochs on the K400 training set and 600 epochs on the UCF101 training set (split-1), with a batch size of 128. The initial learning rate is 0.01 and decreased by the cosine schedule \cite{16arxiv/sgdr}. The optimizer is SGD with a momentum of 0.9 and a weight decay of $1e-4$. After pre-training, the RGB backbone is used as initialization parameters for other video tasks. In ablation study, we conduct all experiments on the subset of K400 with 80k videos like \cite{20eccv/rethinking} and reduce the epochs to 100.

{\noindent\bf Downstream Tasks.} We follow the evaluation protocol in \cite{20nips/coclr}, including two types of downstream tasks. (1) Action classification: we add a single layer for classification and then train the full model with both the linear probe and fine-tune policy. We evaluate the top-1 accuracy. (2) Action retrieval: the backbone is directly used for feature extraction and no further training is required. We take the representations of videos from the test set to query the $k$-nearest neighbours ($k$-NNs) in the training set and report Recall at $k$ (R@k) for comparison.

\begin{table}[!t]
    \centering
    \caption{Ablation study on different designs of MSCL.}
    \begin{tabularx}{1.0\textwidth}{CCCCCCCCC}
        \toprule
        \multicolumn{2}{c}{Contrastive Losses}& \multicolumn{2}{c}{LMCL Policies} & & \multicolumn{2}{c}{Finetune} & \multicolumn{2}{c}{Retrival R@1} \\
        $L_{\mathrm{RF}}$ & $L_{\mathrm{LMC}}$ & FA & MDS & 3D & UCF101 & SSV2 & UCF101 & HMDB51 \\
        \midrule
         & & & & & 65.1 & 39.1 & 22.0 & 13.7\\
        \checkmark & & & & & 71.4 & 41.0 & 36.5 & 22.1\\
        \checkmark & & & & \checkmark & 71.2 & 41.3 & 35.6 & 23.6\\
        & \checkmark & & & & 70.5 & 40.3 & 31.2 & 19.6\\
        \checkmark & \checkmark & & & & 75.6 & 42.2 & 45.7 & 26.7\\
        \checkmark & \checkmark & \checkmark & & & 76.8 & 42.5 & 47.3 & 27.9\\
        \checkmark & \checkmark & & \checkmark & & 76.4 & 42.6 & 46.5 & 27.2\\
        \checkmark & \checkmark & \checkmark & \checkmark & & 77.3 & 42.9 & 48.0 & 28.1\\
        \bottomrule
    \end{tabularx}
    \label{mscl}
\end{table}

\subsection{Ablation Study}
{\bf Motion Sensitive Contrastive Learning.} We analyze how different designs contribute to MSCL, including the contrastive losses $L_{\mathrm{RF}}$ and $L_{\mathrm{LMC}}$ as well as the MDS and FRA strategies. When only $L_{\mathrm{RF}}$ is used, we add the experiment with the 3D backbone in the flow pathway, as there is no need to keep the flow features corresponding to very short-term motions without LMCL. The results are summarized in Table \ref{mscl}. The cross-modality contrastive loss $L_{\mathrm{RF}}$ and local motion contrastive loss $L_{\mathrm{LMC}}$ are complementary. As shown in \cite{21arxiv/modist}, $L_{\mathrm{RF}}$ forces the model to focus on motion areas, which improves the appearance-invariant property (features are activated on motion areas regardless of appearances). Then, $L_{\mathrm{LMC}}$ 
enhances modeling local dynamics, which exhibits consistent performance gains. The comparison also demonstrates the potential of motion information in optical flows, which is not fully explored in recent self-supervised works. MDS and FRA boost the results independently and their combination leads to further improvement. Their contributions lie in different perspectives: MDS provides better training samples for LMCL and FRA aims at motion-related features.

\begin{table}[!t]
    \centering
    \caption{Ablation study on the feature extractor. }
    \begin{tabularx}{1.0\textwidth}{CCCCCCC}
        \toprule
        \multicolumn{2}{c}{Flow Backbone} & & \multicolumn{2}{c}{Finetune} & \multicolumn{2}{c}{Retrieval R@1} \\
        Arch & T-pad & TPN & UCF101 & SSV2 & UCF101 & HMDB51 \\
        \midrule
        R3D-18 & zero & & 71.3 & 41.3 & 35.6 & 23.6\\
        R3D-18 & reflect & & 70.7 & 40.8 & 33.9 & 22.1\\
        ResNet-18 & zero & & 72.3 & 40.8 & 38.6 & 23.9\\
        ResNet-18 & zero & \checkmark & 75.6 & 42.2 & 45.7 & 26.7\\
        \bottomrule
    \end{tabularx}
    \label{feature_extractor}
\end{table}

{\noindent\bf Feature Extractor.} Here, we first study the effect of the 2D backbone in the flow pathway. To achieve this, we exchange 2D ResNet-18 with 3D ResNet-18, whose temporal receptive field is enlarged by pooling and convolution. From the results shown in Table \ref{feature_extractor}, we can see that the 3D network delivers the worst performance, due to that zero padding incurs the leakage of the position information as \cite{20arxiv/position} shows, which degrades LMCL. The result of a 3D network without reflect padding is also inferior, which verifies that detailed dynamics in the flow features encoded by 2D CNNs are more crucial to LMCL. We also present the effect of TPN in the RGB pathway. The model without TPN directly uses the conv3 output and adds one more fully-connected layer for consistency in depth. As Table \ref{feature_extractor} shows, TPN boosts the performance in terms of all the metrics. This can be attributed to the fact that multi-level local features improve the collaboration with global features (from conv5).


{\noindent\bf Interval Sampling.} We study the effect of the proposed MDS by ablating the score function. In Table \ref{sample_methods}, we observe that when combining both the weight and differential maps, the result is significantly improved, much better than either of the single ones. This phenomenon shows that large motion differentials on foreground areas are beneficial for LMCL.

{\noindent\bf Flow Rotation.} Table \ref{rotate_range} shows how different rotation ranges influence video representation learning. From the table, we can see that larger ranges usually lead to better results, which is probably due to that augmented flows increase the difficulty of LMCL, making the features focus more on motion dynamics. It is also noteworthy that the performance decreases a bit when the rotation angle is very small. In this case, the augmented flow is similar to the original one, which confuses the contrastive learning procedure.

{\noindent\bf How LMCL Works?} The model can learn to optimize $L_{\mathrm{LMC}}$ from two aspects: the motion vector itself and the deformation of the object. We show that LMCL indeed takes advantage of both factors. To verify this, we remove $L_{\mathrm{RF}}$ and conduct two kinds of experiments. First, we employ the original $L_{\mathrm{LMC}}$ without augmentation in Eq. \eqref{(5)}, and take the motion boundary (extracted by the Sobel operator \cite{extra/sobel}) as input. In this manner, motion vectors are removed from the flow map. From Table \ref{lmcl_works}, we can see the motion boundary obtains inferior results compared with the flow input, indicating both vector information and deformation are considered in LMCL. Second, we study the superiority of LMCL in learning motion information. To this end, we conduct experiments which directly learn to recognize these transforms. More concretely, we only use the augmented flows as negative samples, \ie, $A(i)$ in Eq. \eqref{(6)} becomes:
\begin{equation}\label{(10)}
   \begin{split}
      & A(i)=\sum_{j=1}^{N}h(\nu^q_{\mathrm{lc}}(i), \hat{m}^q_{j,\textrm{lc}}(i)), \end{split}
\end{equation}
where $\hat{m}^q_{j,\textrm{lc}}(i)$ indicates the local feature of the $j$-th augmented flow at timestamp $i$, and $N$ is the number of augmentations, which is set to 3. Note that, different from Eq. \eqref{(6)}, the augmented negative sample at each timestamp is independent. To take the deformation into consideration, we use shift with the same padding as the extra augmentation. Table \ref{lmcl_works} depicts the results, which, interestingly, show negligible improvement. This, once again, verifies
the necessity and effectiveness of the proposed LMCL.

\begin{figure}[!t]
 \begin{minipage}[t]{0.5\textwidth}
  \centering
     \makeatletter\def\@captype{table}\makeatother\caption{Ablation on sampling methods.}
       \begin{tabular}{cccc} 
	    \toprule
        & \multicolumn{2}{c}{Finetune} \\
        Score Function & UCF101 & SSV2 \\
        \midrule
        - & 75.6 & 42.2 \\
        $\mathrm{sum}(w_t)$ & 75.7 & 42.2 \\
        $\mathrm{sum}(\alpha_t)$ & 76.1 & 42.3 \\
        $\mathrm{sum}(\alpha_t \circ w_t)$ & 76.8 & 42.5 \\
        \bottomrule
	\end{tabular}
	\label{sample_methods}
  \end{minipage}
  \begin{minipage}[t]{0.5\textwidth}
   \centering
        \makeatletter\def\@captype{table}\makeatother\caption{Ablation on rotation ranges.}
         \begin{tabular}{cccc}        
          \toprule
            & \multicolumn{2}{c}{Finetune} \\
            Rotation Range & UCF101 & SSV2 \\
            \midrule
            - & 75.6 & 42.2 \\
            $[\pi/2, 3\pi/2]$ & 76.0 & 42.3 \\
            $[\pi/3, 5\pi/3]$ & 76.8 & 42.5 \\
            $[\pi/6, 11\pi/6]$ & 76.5 & 42.4 \\
            \bottomrule
	  \end{tabular}
	  \label{rotate_range}
   \end{minipage}
\end{figure}


\begin{table}[!t]
    \centering
    \caption{Ablation study on different paradigms for learning motion information.}
    \begin{tabularx}{1.0\textwidth}{cCCCCCC}
        \toprule
        & & & \multicolumn{2}{c}{Finetune} & \multicolumn{2}{c}{Retrival R@1} \\
        Flow Input & $L_{\mathrm{LMC}}$ & Aug. & UCF101 & SSV2 & UCF101 & HMDB51 \\
        \midrule
        flow & Eq.\eqref{(5)} & - & 70.5 & 40.3 & 31.2 & 19.6\\
        boundary & Eq.\eqref{(5)} & - & 68.2 & 39.3 & 30.2 & 17.7\\
        \midrule
        - & - & - & 65.1 & 39.1 & 22.0 & 13.7\\
        flow & Eq.\eqref{(10)} & shift & 66.9 & 39.2 & 27.3 & 15.7\\
        flow & Eq.\eqref{(10)} & rotate & 66.7 & 39.4 & 27.2 & 14.8\\
        flow & Eq.\eqref{(10)} & shift+rotate & 66.9 & 39.5 & 27.2 & 15.7\\
        \bottomrule
    \end{tabularx}
    \label{lmcl_works}
\end{table}

\subsection{Comparison to State-of-the-Art Methods}
\subsubsection{Action Classification.}
We first evaluate our method on the action classification task. 
The results including the linear probe and fine-tune policy are shown in Table \ref{cmp_ac_uh}. In terms of the K400 pre-training setting, MSCL outperforms previous methods with the same R3D-18 backbone. For the major counterpart, MoDist \cite{21arxiv/modist}, MSCL can achieve better top-1 accuracies on both datasets with only half of the epochs. We also notice that the results are still lower than those of the methods like CVRL or $\rho$MoCo and the reason lies in that they use the more advanced R3D-50 backbone 
and more training epochs. When pre-training is applied on UCF101 only, MSCL can achieve better results. We also perform evaluation on the SSv2 dataset in Table \ref{cmp_ac_ssv2} and reproduce MoDist \cite{21arxiv/modist} under the same training setting, where the results show MSCL outperforms others with a 50.3\% top-1 accuracy.

\begin{table}[!t]
    \centering
    \caption{Action classification results on UCF101 and HMDB51. `U+I' denotes the combination of UCF101 and ImageNet. 
    Note that `Sizes' refer to the test setting.
    }
    \begin{tabularx}{1.0\textwidth}{cCCCCCll}
        \toprule
        Method & Network & Year & Dataset & Sizes & Epochs & UCF101 & HMDB51\\
        \midrule
        Playback \cite{20cvpr/videoplayback} & R18 & 2020 & UCF101 & 16$\times$112 & 300 & 69.0/\spaceline   & 33.7/\spaceline \\
        CoCLR \cite{20nips/coclr} & S3D & 2020 & UCF101 & 32$\times$224 & - & 81.4/70.2 & 52.1/39.1\\ 
        MFO \cite{21iccv/mfo} & R18 & 2021 & UCF101 & 16$\times$112 & 300 & 76.2/\spaceline & 44.1/\spaceline \\
        TCLR \cite{21arxiv/tclr} & R18 & 2021 & UCF101 & 16$\times$112 & 400 & 82.4/\spaceline & 52.9/\spaceline \\
        \grey{SeCo \cite{21aaai/seco}} & \grey{R50} & \grey{2021} & \grey{U$+$I} & \grey{-} & \grey{-} & \grey{88.2/\spaceline} & \grey{55.5/\spaceline} \\
        \grey{MCL \cite{21iccv/mcl}} & \grey{S3D} & \grey{2021} & \grey{U$+$I} & \grey{16$\times$224} & \grey{-} & \grey{90.5/79.8} & \grey{63.5/\spaceline} \\
        \midrule
        \textbf{Ours} & R18 & - & UCF101 & 8$\times$112 & 400 & 82.1/72.5 & 53.7/39.9 \\
        \textbf{Ours} & R18 & - & UCF101 & 16$\times$112 & 400 & 86.7/77.1 & 58.9/45.3 \\
        \midrule
        TCLR \cite{21arxiv/tclr} & R18 & 2021 & K400 & 16$\times$112 & 100 & 84.1/\spaceline & 53.6/\spaceline \\
        VideoMoCo\cite{21cvpr/videomoco} & R18 & 2021 & K400 & 32$\times$112 & 200 & 74.1/\spaceline & 43.6/\spaceline \\
        MFO \cite{21iccv/mfo} & R18 & 2021 & K400 & 16$\times$112 & 100 & 79.1/63.2 & 47.6/33.4 \\
        LSFD \cite{21iccv/lsfd} & R18 & 2021 & K400 & 16$\times$112 & 500 & 77.2/- & 53.7/- \\
        ASCNet \cite{21iccv/ascnet}& R18 & 2021 & K400 & 16$\times$112 & 200 & 80.5/- & 52.3/- \\
        MCN \cite{21iccv/mcn}& R18 & 2021 & K400 & 32$\times$128 & 500 & 89.7/73.1 & 59.3/42.9 \\
        TE \cite{21iccv/te}& R18 & 2021 & K400 & 16$\times$128 & 200 & 87.1/- & 63.6/- \\
        MoDist \cite{21arxiv/modist}& R18 & 2021 & K400 & 32$\times$112 & 800 & 91.3/90.4 & 62.1/57.5 \\
        CVRL \cite{21cvpr/cvrl}& R50 & 2021 & K400 & 32$\times$256 & 800 & 92.2/89.2 & 66.7/57.3 \\
        $\rho$MoCo \cite{21cvpr/largescale}& R50 & 2021 & K400 & 8$\times$256 & 200 & 91.0/\spaceline & \spaceline/\spaceline \\
        \midrule
        \textbf{Ours} & R18 & - & K400 & 16$\times$112 & 200 & 90.7/86.1 & 62.3/55.6 \\
        \textbf{Ours} & R18 & - & K400 & 16$\times$112 & 400 & 91.5/88.7 & 62.8/56.5 \\
        
        \bottomrule
    \end{tabularx}
    \label{cmp_ac_uh}
\end{table}
\begin{table}[!t]
    \centering
    \caption{Action classification results on SSv2. $^\dagger$ denotes our reproduced result.}
    \begin{tabularx}{1.0\textwidth}{cCCCCCC}
        \toprule
        Method & Network & Year & Dataset & Size & Epochs & Top-1\\
        \midrule
        RSPNet \cite{21aaai/rspnet} & R18 & 2021 & K400 & 16$\times$112 & 50 & 44.0 \\
        MoDist \cite{21arxiv/modist}$^\dagger$ & R18 & 2021 & K400 & 16$\times$112 & 200 & 49.1 \\
        \textbf{Ours} & R18 & - & K400 & 16$\times$112 & 200 & 50.3 \\
        
        \bottomrule
    \end{tabularx}
    \label{cmp_ac_ssv2}
\end{table}

{\noindent\bf Video Retrieval.}
Similar to the action classification task above, we validate our method with two pre-training datasets and the results are shown in Table \ref{cmp_vr}. On the K400 dataset, our method outperforms the recent state-of-the-art ones with R@1 of 63.7 and 32.6, respectively. On the UCF101 dataset, our method performs better on UCF101 but slightly worse on HMDB51 when compared with the advanced counterpart TE \cite{21iccv/te}. It is worth noting that MCL \cite{21iccv/mcl} applies extra MoCo \cite{20cvpr/mocov1} pre-training on ImageNet, which takes advantage of more training data. These results also demonstrate the effectiveness of MSCL on the retrieval task.

\begin{table}[!t]
    \centering
    \caption{Video retrieval performance on UCF101 and HMDB51. $^\dagger$ indicates additional pretraining on ImageNet is applied.}
    \begin{tabularx}{1.0\textwidth}{ccccCCCCCC}
        \toprule
        & & & & \multicolumn{3}{c}{UCF101} & \multicolumn{3}{c}{HMDB51}\\
        Method & Network & Year & Dataset & R@1 & R@5 & R@10 & R@1 & R@5 & R@10\\
        \midrule
        MemDPC \cite{20eccv/memdpc} & R18 & 2020 & UCF101 & 20.2 & 40.4 & 52.4 & 7.7 & 25.7 & 40.6 \\
        CoCLR \cite{20nips/coclr} & S3D & 2020 & UCF101 & 53.3 & 69.4 & 82.0 & 23.2 & 43.2 & 53.5 \\
        BE \cite{21cvpr/be} & R18 & 2021 & UCF101& 11.9 & 31.3 & 44.5 & - & - & - \\ %
        TCLR \cite{21arxiv/tclr} & R18 & 2021 & UCF101 & 56.2 & 72.2 & 79.0 & 22.8 & 45.4 & 57.8 \\
        MFO \cite{21iccv/mfo} & R18 & 2021 & UCF101 & 39.6 & 57.6 & 69.2 & 18.8 & 39.2& 51.0 \\
        MCN \cite{21iccv/mcn} & R18 & 2021 & UCF101 & 53.8 & 70.2 & 78.3 & 24.1 & 46.8 & 59.7 \\
        TE \cite{21iccv/te} & R18 & 2021 & UCF101 & 63.6 & 79.0 & 84.8 & 32.2 & 61.3 & 71.6 \\
        \grey{MCL \cite{21iccv/mcl}$^\dagger$} & \grey{S3D} & \grey{2021} & \grey{UCF101} & \grey{67.0} & \grey{80.8} & \grey{86.3} & \grey{26.7} & \grey{52.5} & \grey{67.0} \\
        \midrule
        \textbf{Ours} & S3D & - & UCF101 & 63.2 & 78.7 & 83.9 & 25.8 & 52.1 & 66.5 \\
        \textbf{Ours} & R18 & - & UCF101 & 65.6 & 80.3 & 86.1 & 28.9 & 56.2 & 68.3 \\
        \midrule
        SpeedNet \cite{20cvpr/speednet} & S3D & 2020 & K400 & 13.0 & 28.1 & 37.5 & - & - & - \\
        STS \cite{extra/sts} & R18 & 2021 & K400 & 38.3 & 59.9 & 68.9 & 18.0 & 37.2 & 50.7 \\
        MFO \cite{21iccv/mfo} & R18 & 2021 & K400 & 41.5 & 60.6 & 71.2 & 20.7 & 40.8 & 55.2 \\
        LSFD \cite{21iccv/lsfd} & R18 & 2021 & K400 & 44.9 & 64.0 & 73.2 & 26.7 & 54.7 & 66.4 \\
        \midrule
        \textbf{Ours} & R18 & - & K400 & 63.7 & 79.1 & 84.0 & 32.6 & 58.5 & 70.5 \\
        
        \bottomrule
    \end{tabularx}
    \label{cmp_vr}
\end{table}

\section{Conclusion}
In this work, we propose a self-supervised learning framework, namely MSCL, to build motion sensitive video representations.
We perform clip-level contrastive learning with intra-modality and inter-modality losses as well as frame-level contrastive learning LMCL to inject motion dynamics from optical flows into RGB frames. Moreover, FRA and MDS are developed to further enhance the contrastive learning procedure by providing better motion-related features and training samples, respectively. Extensive experiments on standard benchmarks show that our MSCL leads to significant performance gains over state-of-the-art methods in terms of two downstream tasks.

\subsubsection{Acknowledgment.} 
This work is partly supported by the National Natural Science Foundation of China (62022011), the Research Program of State Key Laboratory of Software Development Environment (SKLSDE-2021ZX-04), and the Fundamental Research Funds for the Central Universities.

%
%
\bibliographystyle{splncs04}
\bibliography{macros,egbib}
\end{document}